\documentclass[journal]{IEEEtran}
\usepackage{ifpdf}
\usepackage{authblk}
\usepackage{algorithm}
\usepackage{algpseudocode}
\usepackage{color}
\usepackage{algorithm}
\usepackage{url}
\ifpdf
\usepackage[pdftex]{graphicx}
\else
\usepackage[dvips]{graphicx}
\fi
\usepackage{tabularx}
\DeclareGraphicsExtensions{.eps}
\usepackage[caption=false,font=footnotesize]{subfig}
\usepackage{comment}
\usepackage{bm}
\usepackage{amsmath,amsthm,amsfonts,amssymb}
\usepackage[noadjust]{cite}
\usepackage{nomencl}
\usepackage{tikz}
\usetikzlibrary{automata,positioning}

\theoremstyle{definition}

\begin{document}

\title{A Hybrid PCA-PR-Seq2Seq-Adam-LSTM Framework for Time-Series Power Outage Prediction}
 
\author[1]{Subhabrata Das}
\author[2]{Bodruzzaman Khan}
\author[3]{Xiaoyang Liu}
\affil[1]{Graduate School of Arts and Sciences, Columbia University, New York, USA}
\affil[2]{Sylhet Agricultural University, Bangladesh}
\affil[3]{Department of Electrical Engineering, Columbia University, New York, NY, USA}

\maketitle

\begin{abstract}
Accurately forecasting power outages is a complex task influenced by diverse factors such as weather conditions \cite{8656482}, vegetation, wildlife, and load fluctuations. These factors introduce substantial variability and noise into outage data, making reliable prediction challenging. Long Short-Term Memory (LSTM) networks, a type of Recurrent Neural Network (RNN), are particularly effective for modeling nonlinear and dynamic time-series data, with proven applications in stock price forecasting \cite{li2019dp}, energy demand prediction, demand response \cite{shi2021end}, and traffic flow management \cite{zhu2016traffic}. This paper introduces a hybrid deep learning framework, termed PCA-PR-Seq2Seq-Adam-LSTM, that integrates Principal Component Analysis (PCA), Poisson Regression (PR), a Sequence-to-Sequence (Seq2Seq) architecture, and an Adam-optimized LSTM. PCA is employed to reduce dimensionality and stabilize data variance, while Poisson Regression effectively models discrete outage events. The Seq2Seq-Adam-LSTM component enhances temporal feature learning through efficient gradient optimization and long-term dependency capture. The framework is evaluated using real-world outage records from Michigan, and results indicate that the proposed approach significantly improves forecasting accuracy and robustness compared to existing methods.
\end{abstract}


\IEEEpeerreviewmaketitle

\graphicspath{}

\section{Introduction}

In modern societies, which are heavily reliant on electricity \cite{:/content/journals/10.1049/ip-a-3.1991.0001}, the economic impacts of power outages have been escalating in certain regions due to aging infrastructure and shifting weather patterns \cite{caswell2010weather}. Severe weather events, in particular, are among the most frequent natural phenomena that can cause widespread power disruptions, costing the U.S. economy billions of dollars annually \cite{smith2016us} and affecting millions of utility customers \cite{changeblackout}. Utilities can reduce disruption during planned outages through careful crew and operation management. However, unplanned outages from system faults often cause extended interruptions, making their reduction and management a top priority \cite{rustebakke1983electric, jaech2018real}. Despite advancements in electricity network design, outages remain inevitable, driving significant research interest in outage prediction  \cite{article_Negative_binomial}, \cite{article_outage_hurricane}.

In recent years, machine learning algorithms have been increasingly employed to address various challenges in power grid management, including forecasting, risk analysis \cite{guikema2009natural}, fault identification in distribution systems \cite{thukaram2005artificial}, security assessment \cite{621229}, and predicting the duration of power outages \cite{nateghi2011comparison}. Over the past two decades, the use of machine learning for predicting power disruptions has become a standard practice \cite{yang2021effect}, \cite{nateghi2014power}, \cite{he2017nonparametric}, \cite{cerrai2019predicting}, \cite{kabir2019predicting}. These methods have demonstrated their effectiveness in enhancing grid reliability and operational efficiency.

This study makes three key contributions: (a) The development of a hybrid model that integrates concepts from Sequence-to-Sequence (Seq2Seq) modeling, Long Short-Term Memory (LSTM) networks, Poisson distribution, and activation functions. (b) The training and optimization of this hybrid model using real-world weather-related event data, with the goal of predicting outage frequencies across different regions of Michigan State. (c) The benchmarking of the proposed methodology against other approaches using actual outage data, providing a robust evaluation of its predictive performance. Through these contributions, this research aims to advance the field of outage prediction and support more effective grid management strategies.

\section{Background}

Effectively mitigating unplanned outages requires a clear understanding of their frequency, duration, and affected consumers. However, their unpredictability poses a challenge \cite{kankanala2013adaboost}, driven by complex factors such as weather conditions \cite{article_cost_of_poor_power_quality}, \cite{climate_change}. Early power outage prediction relied on statistical methods like generalized linear models \cite{li2010statistical} and mixed models \cite{guikema2006statistical}, primarily focusing on specific storms like hurricanes and ice storms \cite{liu2007statistical}. Later, hybrid approaches emerged, including ensemble models for outage counts \cite{guikema2012hybrid} and adaptable decision tree ensembles for various storm types \cite{wanik2015storm}, \cite{alpay2020dynamic}.

Most studies now use statistical and machine learning methods to predict outage frequency, focusing on weather features, especially during extreme events like hurricanes, thunderstorms, and snowstorms \cite{zhu2007storm}, \cite{liu2008spatial}, \cite{alvehag2010r}, \cite{jaech2018real}. Eskandarpour and Khodaei \cite{eskandarpour2016machine} utilized logistic regression to predict outages caused by an approaching hurricane, while Nateghi et al. \cite{nateghi2014power} applied random forest (RF) to forecast outages resulting from tropical cyclones. He et al. \cite{he2017nonparametric} and Cerrai et al. \cite{cerrai2019predicting} employed Bayesian additive regression trees (BART) for storm-related outage predictions. Kabir et al. \cite{kabir2019predicting} used RF, support vector machine (SVM), boosting tree (BT), and quantile regression forest (QRF) to predict power outages caused by thunderstorms. Liu et al. \cite{liu2008spatial} applied Generalized Linear Models (GLMs) to analyze the significance of hurricane and ice storm variables in outage prediction, refining the earlier hurricane outage prediction model by \cite{liu2005negative}.

While these models have shown promising results, further improvements should consider factors beyond weather alone. Current statistics indicate that many routine outages, affecting hundreds of customers, are caused by tree branches falling onto overhead distribution lines \cite{article_reliability_tree}. Several studies have estimated the average number of outages under normal conditions \cite{domijan2005effects,kankanala2012estimation}, while others assessed the susceptibility of power system components to failure using machine learning \cite{rudin2011machine, gross2006predicting, jaech2018real}. To improve outage frequency prediction, statistical methods have been employed to quantify feature-outage relationships. Chow et al. \cite{chow1996time} analyzed the statistical significance of various factors affecting outage duration but did not propose a forecasting model. Adibi and Milanicz \cite{adibi1999estimating} introduced an estimation approach based on the restoration process, which requires detailed insights into outage characteristics and recovery steps. Rodriguez and Vargas \cite{rodriguez2005fuzzy} developed a fuzzy logic method that reduces reliance on detailed repair data but depends on expert input.

With the evolution of deep learning \cite{lecun2015deep}, researchers are developing new frameworks for optimization and reliable predictions. Current trends focus on optimizing models \cite{johnston2023curriculum}, making them lightweight \cite{10720908}, and computationally inexpensive by using approaches like Bayesian optimized hybrid learning \cite{khan2024bayesian} and perturbed decision-focused learning \cite{yi2024perturbed}. There is also growing interest in multi-purpose machine learning and novel deployments, such as federated learning-based building-level load forecasting \cite{10063999} and the implementation of edge intelligence in smart meters \cite{li2024introducing}.

In the context of deep learning for outage prediction, relatively few comprehensive studies exist. However, due to its remarkable nonlinear learning capabilities and excellent prediction performance \cite{KHAN2018241_deeplearning_application}, coupled with advancements in computational power that have revolutionized forecasting, Convolutional LSTM (ConvLSTM) \cite{10138051} has emerged as an end-to-end trainable model for precipitation forecasting. This model builds upon the fully connected LSTM (FC-LSTM) by incorporating convolutional structures into both the input-to-state and state-to-state transitions \cite{shi2015convolutional}.

In this context, we aim to explore the application of deep neural networks to our outage prediction problem and evaluate their performance in comparison to traditional machine learning approaches.

\begin{figure*}
\centering
           \includegraphics[width=1\textwidth]{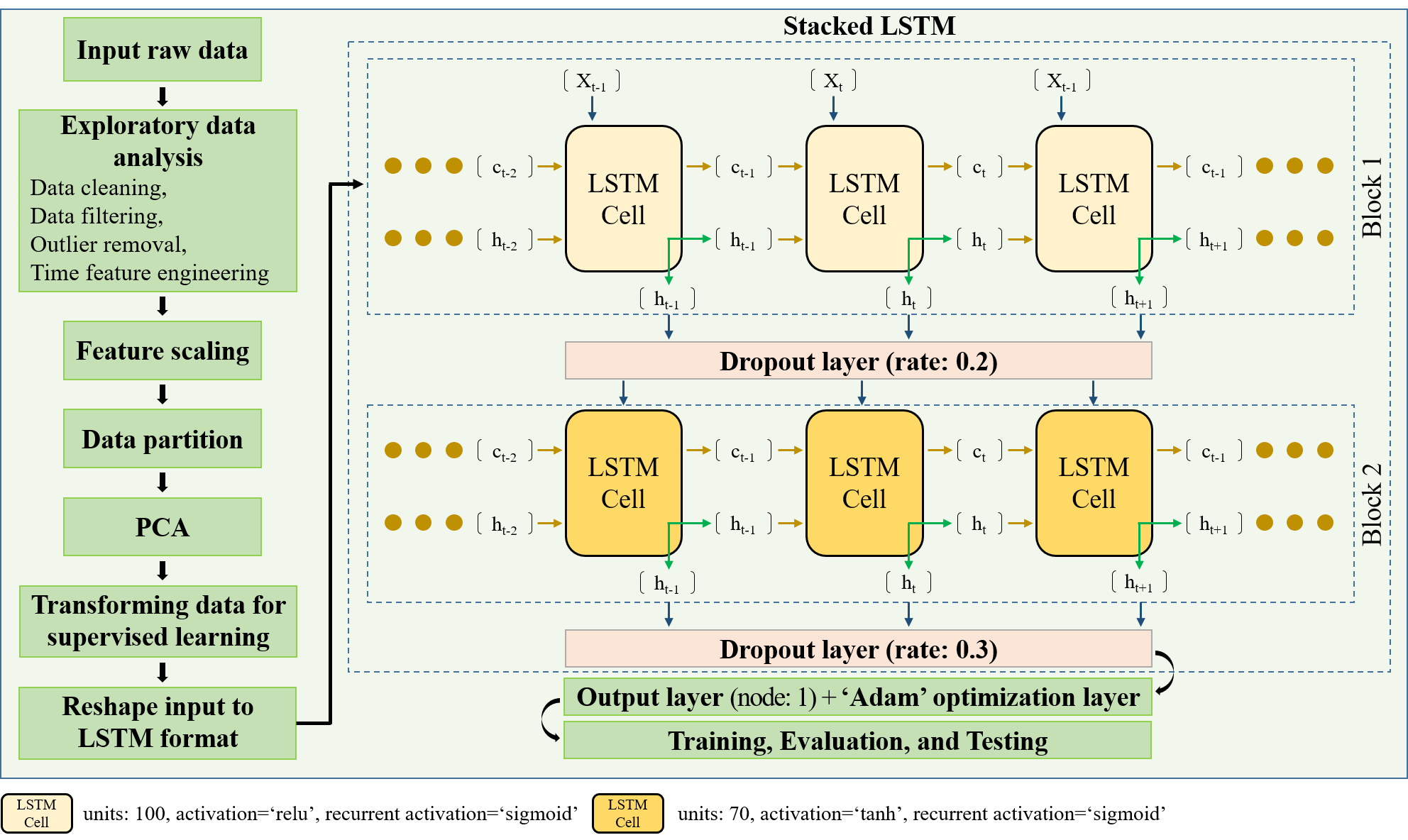}
           \caption{Schematic and process-flow of the proposed Hybrid (PCA-PR-Seq2Seq-Adam-LSTM) Model.}
             \label{Schematic}	     	     
\end{figure*}

\section{Problem Statement}

The objective is to predict the frequency of power outages at specific locations over time, leveraging a hybrid Long Short-Term Memory (LSTM) model with sequence-to-sequence (seq-to-seq) architecture. The predictions are time- and location-specific, focusing on modeling outage frequencies as discrete occurrences that follow a Poisson distribution.

The power outage prediction model using a hybrid LSTM framework represents a significant innovation in forecasting grid reliability and operational resilience. By integrating sequence-to-sequence learning with Poisson optimization, the model captures the complex temporal dependencies and stochastic nature of power outages more effectively than traditional methods. This dual-layered approach allows the model to process time-series data and predict discrete event counts, such as the number of outages, with reasonably high accuracy. The incorporation of both temporal and spatial dimensions ensures predictions are location- and time-specific, providing utilities with actionable insights to optimize resource allocation and enhance grid stability. Furthermore, the hybrid LSTM model's ability to adapt to non-linear patterns in the data and account for external influences, such as weather conditions and demand surges, makes it a cutting-edge solution for proactive outage management. This innovation not only improves prediction precision but also fosters sustainable energy practices by enabling preventive measures and reducing downtime.

\subsection{Mathematical Formulation}

\subsubsection{Input Features} Let $X={x_1,x_2...x_T}$ represent the time series data for $T$ time steps.

Each $x_t\in\mathbb{R}^d$ contains d-dimensional features such as: weather data (temperature, wind speed, precipitation, etc.), historical power outage data.

\subsubsection{Target Output} $Y={y_{t+1},y_{t+2},...,y_{t+H}}$ is the target sequence, where $y_{t+h}$ represents the predicted outage frequency at time $t+h$ for a specific location. Each $y_{t+H}$ is a non-negative integer, $y_{t+H}\in N_0$, and is modeled using a Poisson distribution.

\subsubsection{Seq-to-Seq Model Architecture}

Encoder:
Input sequence $X$ is passed through an LSTM encoder to capture temporal dependencies and generate a context vector $h_t=LSTM_{e_{nc}}(x_t,h_{t-1})$ for $t=1,...,T$.
$C = f_{enc}(h_1, h_2, \dots, h_T)$

Decoder:
The context vector $C$ initializes the LSTM decoder to predict the sequence $Y$ iteratively.
$\hat{y}_{t+h} = f_{dec}(C, h_{t+h-1})$
for $h=1,...,H$

\subsubsection{Poisson Optimization}

The outage frequency is modeled as a Poisson random variable with rate parameter $\lambda_{t+h}$, where $\lambda_{t+h}$ is predicted by the LSTM decoder.
$\hat{\lambda_{t+h}}=exp(W_{\lambda} h_{t+h} + b_{\lambda})$
The likelihood of observing $y_{t+h}$ is: $P(y_{t+h}|\hat{\lambda_{t+h}})=\hat{\lambda}_{t+h}^{y_{t+h}}exp(-\hat{\lambda_{t+h}})/(y_{t+h})!$

\subsubsection{Loss Function}

The loss function is the negative log-likelihood for the Poisson distribution:
\begin{equation}
L_{Poisson}=-\sum_{h=1}^{H}\sum_{n=1}^{N} [y_{t+h}^nlog(\hat{\lambda}_{t+h}^n)-\hat{\lambda}_{t+h}^n-log(y_{t+h}^n!)]
\end{equation}





\section{Methodology}


\begin{figure*}
\centering
           \includegraphics[width=1\textwidth]{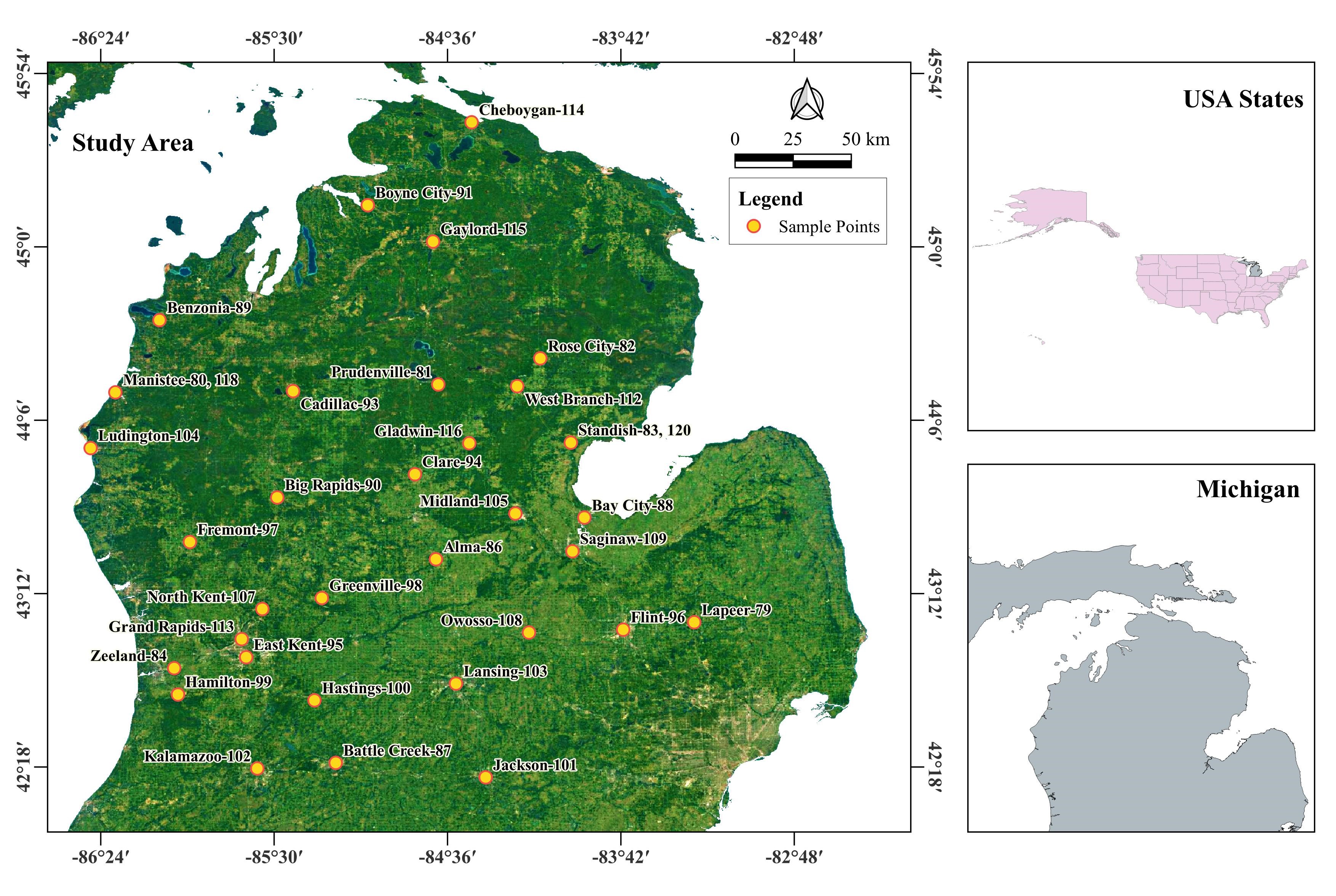}
           \caption{Study Area Map}
             \label{Schematic}	     	     
\end{figure*}

\subsection{Spatial Autocorrelation}
Before introducing the proposed methodology, we investigated the spatial autocorrelation of the actual outage data. 


Utilizing Moran's I test \cite{moran1950notes}, a widely used test for spatial autocorrelation, we examined the following hypotheses.

\begin{itemize}
    \item \textbf{Null Hypothesis (H$_0$):} Outages are spatially random, with no autocorrelation.
    \item \textbf{Alternative Hypothesis (H$_1$):} Outages show significant spatial autocorrelation.
\end{itemize}

The mathematical equation employed for testing Moran's I is given by:

\[
I = \frac{N}{W} \cdot \frac{\sum_{i=1}^{N} \sum_{j=1}^{N} w_{ij} (x_i - \bar{x})(x_j - \bar{x})}{\sum_{i=1}^{N} (x_i - \bar{x})^2}
\]

Where
\begin{itemize}

    \item \(x_i\) and \(x_j\) are the attribute values at spatial units \(i\) and \(j\).
    \item \(w_{ij}\) is the spatial weight between units \(i\) and \(j\).
    \item \(N\) is the total number of spatial units.

    \item \(\bar{x}\) is the mean of all the values \(x_i\).
    \item \(W\) is the sum of all the weights in the spatial weight matrix: \(W = \sum_{i=1}^{N} \sum_{j=1}^{N} w_{ij}\).
\end{itemize}

The Moran's I value of -0.0383, along with a p-value of 0.132 and a z-score of -0.2557, suggests that there is no statistically significant geographic autocorrelation in the distribution of outages. The near-zero Moran's I indicates spatial distribution of outages lacks a distinct clustering or dispersion pattern. As a result, there is no strong evidence that outages vary spatially within the research area.

\begin{table}[h!]
    \captionsetup{justification=raggedright, format=plain, singlelinecheck=false} 
    \caption{ Moran's I Test Results for Spatial Autocorrelation of Outages}
    \renewcommand{\arraystretch}{1.5} 
    \begin{tabular}{|p{4cm}|p{4cm}|} 
        \hline
        \textbf{Statistic} & \textbf{Value} \\ 
        \hline
        Moran's I          & -0.0383        \\
        p-value            & 0.132          \\
        z-score            & -0.2557        \\
        \hline
    \end{tabular}
\end{table}

\subsection{Proposed PCA-PR-Seq2Seq-Adam-LSTM Model}

The construction of a hybrid (PCA-PR-Seq2Seq-Adam-LSTM) outage model includes the systematic incorporation of the model elements from the preprocessing state into the final optimization layer. In this work, we have implemented a sequence-to-sequence model architecture using the LSTM network. LSTM is very suitable for handling time series and sequential data. Further optimization is done with the Adam optimizer, which is known for its adaptive learning rate and computational efficiency. Before feeding to the model, we preprocessed the data using PCA in order to make the data well-structured for the model so that it improves the performance of the models in later stages. We have also used Poisson regression to model count outcomes emanating from our dataset, which suited the target variables quite well. The study investigated various activation functions: the rectified linear unit (ReLU), sigmoid, and leaky ReLU. All these tests were conducted with the purpose of testing the influence of different methods of activation on convergence and the precision of the model. The detailed description of these combinations is described below.
The pipeline of the proposed approach is shown in algorithm 1.

\subsubsection{PCA}

 In the PCA algorithm, orthogonal transformation was used to convert a set of weather variables by virtue of dimensionality reduction. The converted variables are used as the principal components.

\subsubsection{Seq2seq}
Earlier, LSTM was able to address the issue of long-term dependencies, but it had a drawback—it required the input and output to have the same number of time steps \cite{xiang2020rainfall}. To overcome this, \cite{cho2014learning} introduced a neural network architecture known as the Encoder-Decoder, or seq2seq, which allows for different input and output time step lengths.

A standard Seq2Seq model consists of an encoder that transforms an input sequence $x = (x_{1}, . . . , x_{T})$ into an intermediate representation $h$, and a decoder that generates the output sequence $y = (y_{1}, . . . , y_{K})$ based on $h$ \cite{sutskever2014sequence}. Additionally, the decoder can leverage an attention mechanism to focus on specific parts of the encoder's states.


\subsubsection{Poisson Regression}

Another property we noticed from the nature of our data is that the outage incident number obeys the Poisson distribution under certain parametric conditions. Thus, we would like to find out the mean of each Poisson distribution under a certain condition, the combination of weather and other features. 
$f(k;\lambda)=Pr(X=k)=\frac{\lambda ^k e^{-\lambda}}{k!}$
where
    - $k$ is the number of outage incident number ($k = 0, 1, 2...)$
    - $e$ is Euler's number.

\subsubsection{Data Denoising}
Other factors, such as the measurement errors may introduce some noise obeying the normal/Gaussian distribution. Thus, we processed the data fed into the LSTM Model by introducing the denoising procedure. 

$f(x)=\frac{1}{\sigma \sqrt{2\pi}} e^{-\frac{(x-\mu)^2}{2 \sigma ^2}}$

where
    - $x$ is the feature tensor
    - $\mu$ is the mean
    - $\sigma$ is the standard deviation\\



\subsubsection{Proposed Stacked LSTM Configuration}

In this study, we propose a multi-step forecasting model using an Encoder-Decoder LSTM architecture with stacked hidden LSTM layers. The model takes multivariate input data and outputs a vector with 1-7 elements, one for each day in the output sequence. We tuned the architectural components of the LSTM, such as the number of hidden layers and nodes per layer, using a grid search. Prior to feeding the input data into the custom LSTM, we normalized it to ensure consistent scale and improve model convergence.

Our proposed LSTM architecture is divided into two blocks, each designed to capture outage-related patterns with specialized configurations. Block 1 consists of 100 cells with ReLU activation, while Block 2 contains 70 cells with tanh activation. This hierarchical design enables improved feature extraction and better adaptability to complex temporal patterns. To effectively manage the flow of information across the network, we introduce specialized gating mechanisms within each block, which regulate the retention and updating of the cell state.

The forget gate regulates the retention of information from the previous cell state ($c_{t-1}$), defined as:

\[
f_t = \sigma(W_f \cdot [h_{t-1}, x_t] + b_f)
\]

where $W_f$ and $b_f$ are the weight matrix and bias vector, and $\sigma$ is the sigmoid activation function.

The input gate updates the cell state ($c_t$) by incorporating new information. The candidate cell state ($\tilde{c}_t$) is calculated differently for each block:

\[
\tilde{c}_t = \text{ReLU}(W_c \cdot [h_{t-1}, x_t] + b_c) \quad \text{(Block 1)}
\]
\[
\tilde{c}_t = \tanh(W_c \cdot [h_{t-1}, x_t] + b_c) \quad \text{(Block 2)}
\]

The input gate activation and updated cell state are:

\[
i_t = \sigma(W_i \cdot [h_{t-1}, x_t] + b_i)
\]
\[
c_t = f_t \otimes c_{t-1} + i_t \otimes \tilde{c}_t
\]

where $W_c$, $W_i$, $b_c$, and $b_i$ are the respective weights and biases.

The output gate determines the hidden state ($h_t$), which propagates temporal features to subsequent timesteps. The output gate activation and hidden state computation differ across blocks:

\[
o_t = \sigma(W_o \cdot [h_{t-1}, x_t] + b_o)
\]
\[
h_t = o_t \otimes \text{ReLU}(c_t) \quad \text{(Block 1)}
\]
\[
h_t = o_t \otimes \tanh(c_t) \quad \text{(Block 2)}
\]

This dual-block architecture improves upon traditional LSTM designs by incorporating distinct activation mechanisms, enhancing feature extraction and enabling more dynamic information flow for learning complex temporal dependencies.


\begin{algorithm}[htbp]
\caption{Pipeline of the Proposed Framework}
\label{alg:comprehensive_pipeline}
\begin{algorithmic}[1]

\Require 
\begin{itemize}
    \item \textbf{Input:} Time series data 
    $\mathbf{X}$, model parameters, input features $\mathbf{F}$, target variable $\mathbf{Y}$
\end{itemize}

\Ensure
\begin{itemize}
    \item \textbf{Output:} Processed data, trained model, forecasted values
\end{itemize}  

\hspace{-4em} \textbf{Step 1: Data Preprocessing}
\begin{enumerate}
    \item Check Spatial Autocorrelation by Moran’s I test
    \item Compute aggregate statistics: $\mathbf{X}_{\text{agg}} = \text{GroupBy}(\text{date, region})$
    \item Merge aggregates: $\mathbf{X} \leftarrow \mathbf{X} \cup \mathbf{X}_{\text{agg}}$
    \item Define bounds: $Q_1 = \text{quantile}(\mathbf{X}, 0.25)$, $Q_3 = \text{quantile}(\mathbf{X}, 0.99)$
    \item Filter data: $\mathbf{X} = \{x \in \mathbf{X} \mid Q_1 \leq x \leq Q_3\}$
    \item Add temporal features: $\{\text{day}, \text{month}, \text{season}\}$
\end{enumerate}

\hspace{-4em} \textbf{Step 2: Convert Time Series to Supervised Format}
\begin{enumerate}
    \item Lagged inputs: $\mathbf{X}_{\text{lag}} = \{\mathbf{X}(t-i) \mid i = 1, \dots, n_{\text{in}}\}$
    \item Forecast outputs: $\mathbf{Y}_{\text{forecast}} = \{\mathbf{Y}(t+j) \mid j = 0, \dots, n_{\text{out}}\}$
\end{enumerate}

\hspace{-4em} \textbf{Step 3: Data Preparation for Model Training}
\begin{enumerate}
    \item Normalize: $\mathbf{X} \leftarrow \text{MinMaxScale}(\mathbf{X})$
    \item Reshape to supervised format: $\mathbf{X}_{\text{supervised}}$
    \item Split into train/test: $\mathbf{X}_{\text{train}}, \mathbf{X}_{\text{test}}$
    \item Dimensionality reduction (PCA): $\mathbf{X} \leftarrow \text{PCA}(\mathbf{X})$
\end{enumerate}

\hspace{-4em} \textbf{Step 4: Proposed LSTM Model Setup}
\begin{enumerate}
    \item Define architecture: $\text{Stacked-LSTM}(L, N, D)$, where $L =$ layers, $N =$ neurons, $D =$ dropout
    \item Compile with loss and optimizer
    \item Train on $\mathbf{X}_{\text{train}}$, apply early stopping
    \item Evaluate on $\mathbf{X}_{\text{test}}$ with metrics (RMSE, MAPE, SMAPE, $R^2$)
\end{enumerate}

\hspace{-4em} \textbf{Step 5: Forecast Generation}
\begin{enumerate}
    \item Preprocess forecast data: scale and transform
    \item Reshape forecast data: supervised format
    \item Predict: $\hat{\mathbf{Y}} = \text{Stacked-LSTM}(\mathbf{X}_{\text{forecast}})$
    \item Inverse scaling on $\hat{\mathbf{Y}}$
    \item Calculate forecast metrics
\end{enumerate}

Processed data, trained model, forecasted values
\end{algorithmic}
\end{algorithm}

\subsubsection{Dropout Layers} 
To avoid overfitting and get a sparse structure, a dropout rate of 0.2 and 0.3 were used to randomly delete 20$\%$ and 30$\%$ of neural connections between subsequent layers during training. Through this operation, dropout layers function as regularizers, allowing the model to be generalized for unknown inputs. \\


All dense layers were activated using the ReLU activation, and the hybrid model was compiled using the Adam optimizer.

\section{Case Study}

\subsection{Study Area}
The study was conducted in different regions of Michigan State. The dataset includes many planning plants and Outage Management System (OMS), each with its own unique region code. Some key planning plants are as follows: Lapeer-79, Manistee-80, Prudenville-81, Rose City-82, Standish-83, and Zeeland-84. Besides, some of the OMS regions include Adrian-85, Alma-86, Battle Creek-87, Bay City-88, Benzonia-89, and Big Rapids-90. The outage dataset consists of real-time customer-reported outage occurrences organized by its OMS, which is also used to generate service tickets for directing repair crews. We define an outage as damage to a component of the power grid. 

The graphical representation of the study location with the specific sites is illustrated in Figure 1.

\subsection{Data Description}

This dataset provides a comprehensive overview of weather-related events for the state of Michigan, capturing multiple meteorological variables that influence local conditions and weather patterns. Key data points include wind speed and wind gust, which measure the average and peak strength of wind, respectively, providing insights into storm potential and energy distribution. Wind-bearing speed further details the directionality of the winds, indicating weather fronts and potential storm paths.

Cloud cover percentages reveal the extent of sky occlusion, essential for understanding sunlight availability, temperature regulation, and precipitation likelihood. Snow cover data highlights seasonal accumulation and its variability, which is vital for tracking snowmelt, runoff, and potential flood risks. Instances of thunderstorms are recorded to assess severe weather conditions, as these events can lead to lightning, hail, and high winds that impact infrastructure. Similarly, rainfall data quantifies precipitation, essential for analyzing patterns related to agricultural planning, water resources, and flood prediction.

These variables provide a robust framework to examine and model weather events across Michigan, enabling better forecasting, risk assessment, and understanding of climate trends within the region.





\subsection{Evaluation Metrics}
We have employed a wide range of evaluation metrics, such as Mean Absolute Error (MAE), Mean Absolute Percentage Error (MAPE), Symmetric Mean Absolute Percentage Error(SMAPE), Median Absolute Percentage Error(MdAPE), and Root Mean Squared Error (RMSE) to demonstrate the robustness of our proposed hybrid model.

\begin{align}
    MAE = \frac{1}{n}\sum_{i=1}^{n} \left|\hat{x_{i}}-x_{i}\right|= p_{t}
\end{align}
\begin{align}
    MAPE = \frac{1}{n}\sum_{i=1}^{n} \frac{\left|x_{i}-\hat{x_{i}}\right|}{x_{i}}
\end{align}
\begin{align}
    SMAPE = \frac{2}{n}\sum_{i=1}^{n} \left|\frac{x_{i}-\hat{x_{i}}}{x_{i}+\hat{x_{i}}}\right|
\end{align}
\begin{align}
    MdAPE = median(p_{1}/x_{1},p_{2}/x_{2},...,p_{N}/x_{N})
\end{align}
\begin{align}
    RMSE = \sqrt{\frac{1}{n}\sum_{i=1}^{n} (x_{i}-\hat{x_{i}})^2}
\end{align}\

\subsection{Prediction accuracy metrics}

Figure \ref{Feb24predictionmetrics} shows the model prediction metrics for February 23, 2021, across the regions. Based on the error rates indicated through the use of RMSE, the range is between a low of 2.79 and a high of 19.02, while Region Code 89 has the lowest error rate among the regions. Thus, this region has very high accuracy. In relation to MAPE, the values vary between 72.15 and 152.62, while the same region provides the minimum value. 

When evaluating SMAPE, the scores range from 50.18 to 66.36, with the best score of the lowest error for Region Code 112. Also, the performance of the model varies for $R^2$, with the highest value achieved for Region Code 112. 

Overall, the performances of the model yield an average mean RMSE of 7.89, a mean MAPE of 117.26, and a mean SMAPE of 60.65 across all regions, with a standard deviation of 4.22, 19.52, and 4.76, respectively. 

\subsection{Comparison to the state-of-the-art models}

In the proposed architecture, two kinds of loss functions were used in order to optimize the prediction accuracy. First, we applied Poisson loss, particularly fitted to count data since it accounts inherently for the properties of the Poisson distribution in event-based forecasting. In case the predictions result in some negative values, the model defaults to MSE. The dual approach provides robust model performance against the observed data. Considereing these loss functions for a specific region, Figure  \ref{loss_} has been designed where training loss values range from 0.3551 to 0.3557, while validation loss values range between 0.3069 and 0.3075. These loss values are consistent, therefore suggesting stability in model training and minimal overfitting with good generalization on the validation dataset. On average, the model obtained a training loss of 0.3553 and a validation loss of 0.307. While these values are moderately higher than the existing outage models \cite{hou2020data, fatima2024machine} MSE (0.161, 0.114, 0.104, 0.061, and 0.059 for Ridge, SVR, CART, GBDT, and RF, respectively), the increase in our study is due to real-world data with broader variability and complexity.

\subsection{Outage Forecasts}
\begin{figure}
\centering
           \includegraphics[width=0.5\textwidth]{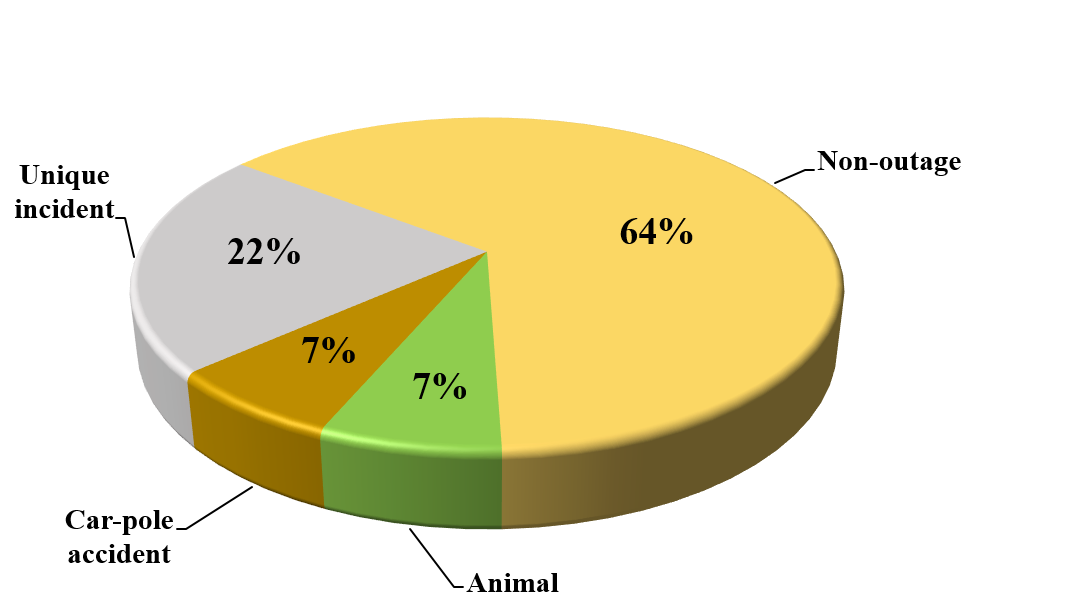}
           \caption{Count of non-weather related  events for region 112 on Feb 23, 2021}
             \label{Incidentbreakdown2}	     	     
\end{figure}

\begin{figure}
\centering
           \includegraphics[width=0.5\textwidth]{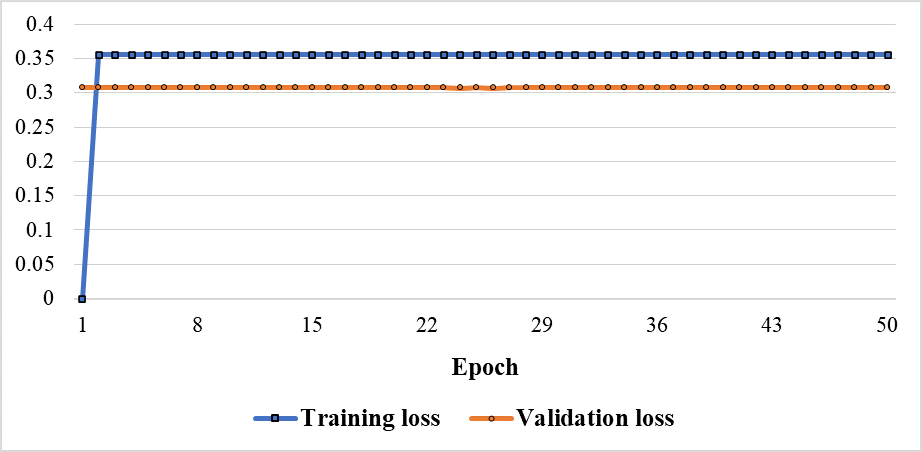}
           \caption{Line plot of training and validation loss over 50 epochs }
             \label{loss_}	     	     
\end{figure}

\begin{figure}
\centering
           \includegraphics[width=0.5\textwidth]{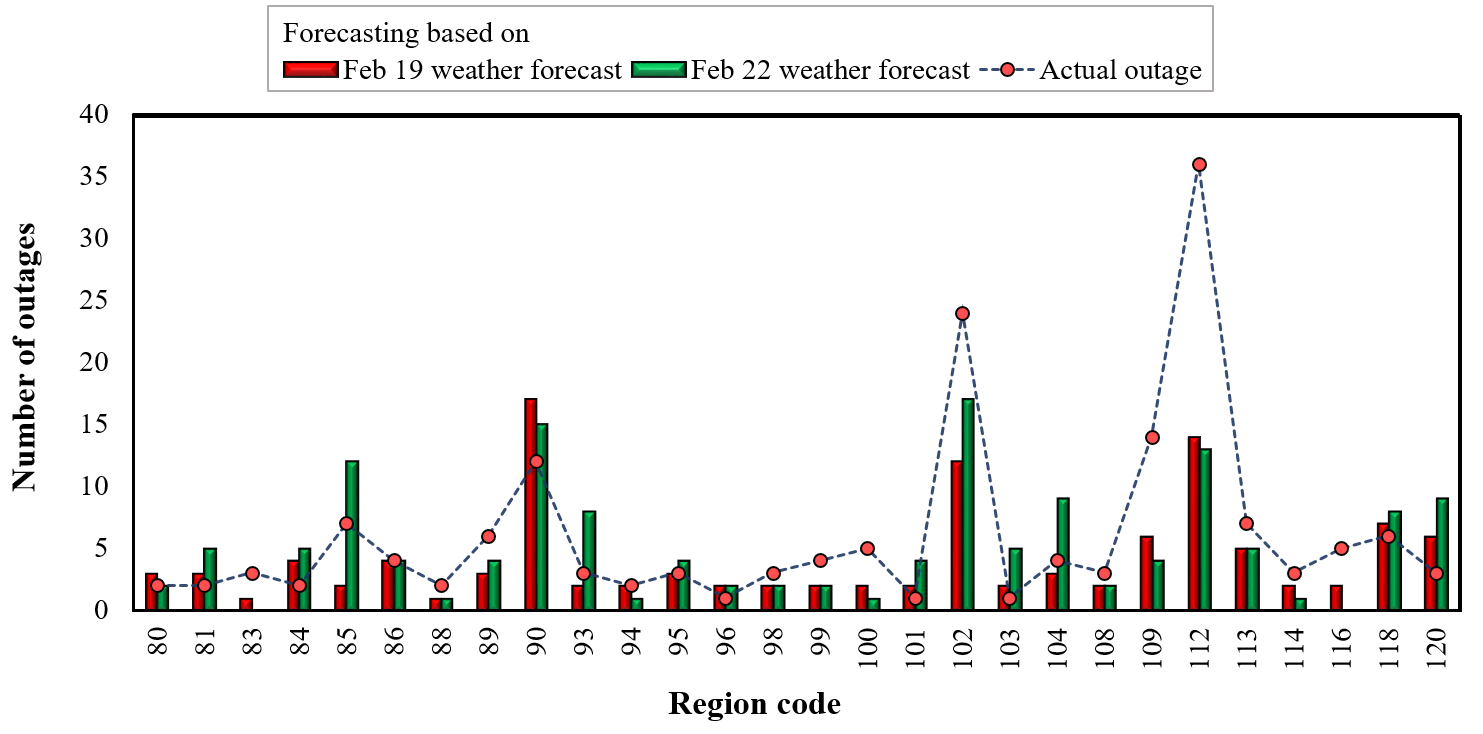}
           \caption{Outage Predictions of Feb 23, 2021}
             \label{Feb23Outage}	     	     
\end{figure}

The figure \ref{Feb23Outage} compares the forecast of power outages using two different weather predictions—one from February 19 and one from February 22, 2021—against the actual outages that occurred on February 23 within different regions. The forecasts show variability in accuracy, as some regions, such as 80 and 86, demonstrate very close alignment between predicted and actual outages. More specifically, Region 80 was forecasted for 2 outages based on data from February 22 and is consistent with the actual outage of 2; Region 86 had a forecast of 4 and actual of 4. However, regions such as 102 and 112 show significant discrepancies whereby actual outages were much larger than forecasted. The disparity can be explained by considering other crucial factors. For Region 112, the discrepancy between forecasted and actual outages was partly due to non-weather causes. The forecasts project that the outages, based on weather predictions of February 22 and 19, would reach 13 and 14, respectively, but actually reached a peak of 36; this foregoing projection was generated by animal interference from 7\%, car-pole accident exposure from 7\%, unique incidents from 22\%, and non-outage factor from 64\%—all non-weather events (figure \ref{Incidentbreakdown2}). These have contributed to the higher-than-expected outage.

Overall, the forecast depicts more accurate results for the more recent weather data because the 145 total outage forecasted on February 22 is closer to 156 actual than the 116 forecasted on February 19.

\subsection{Forecast Accuracy Comparison}
\begin{figure}
\centering
           \includegraphics[width=0.5\textwidth]{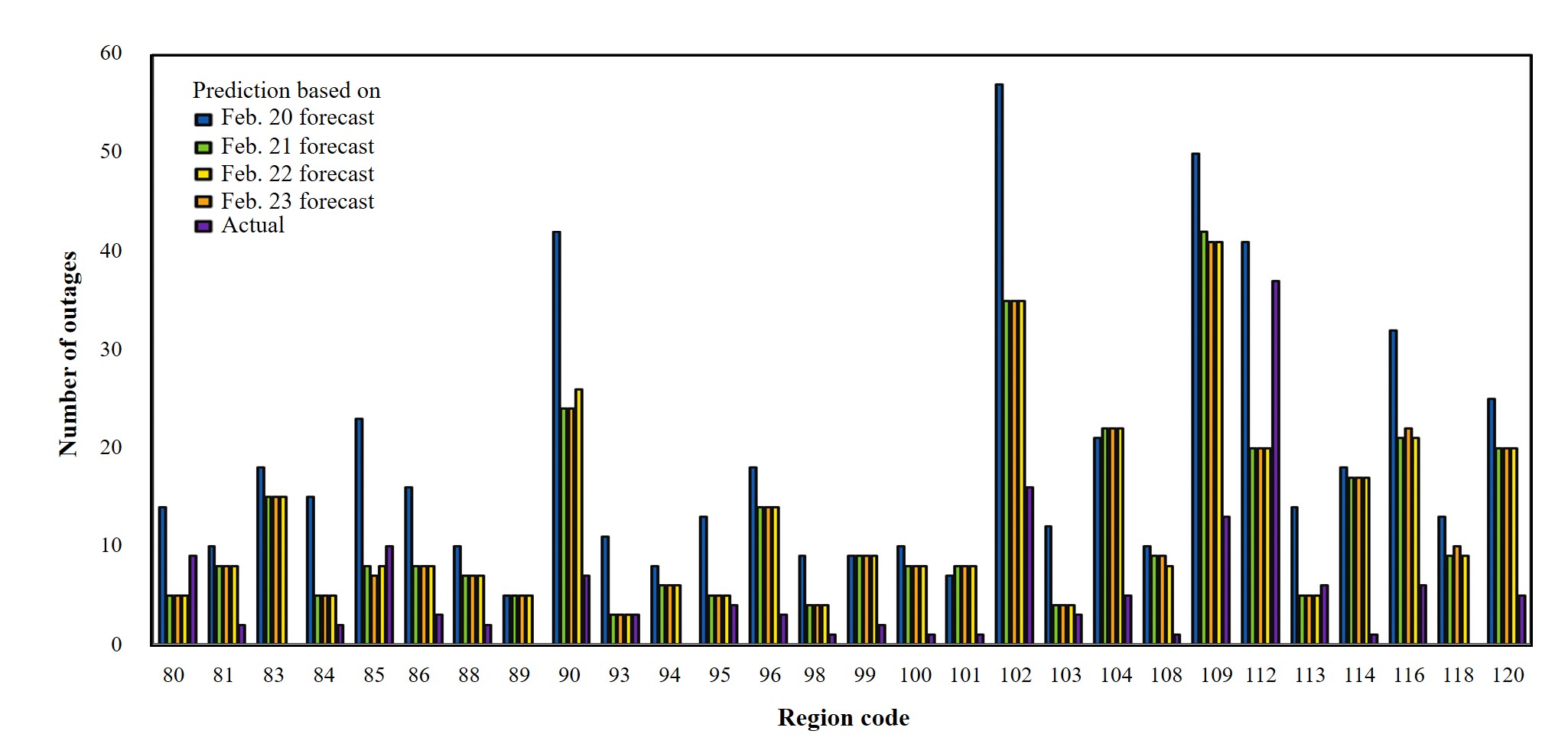}
           \caption{Outage Predictions of Feb 24, 2021}
             \label{Feb24prediction}	     	     
\end{figure}

\begin{figure}
\centering
           \includegraphics[width=0.5\textwidth]{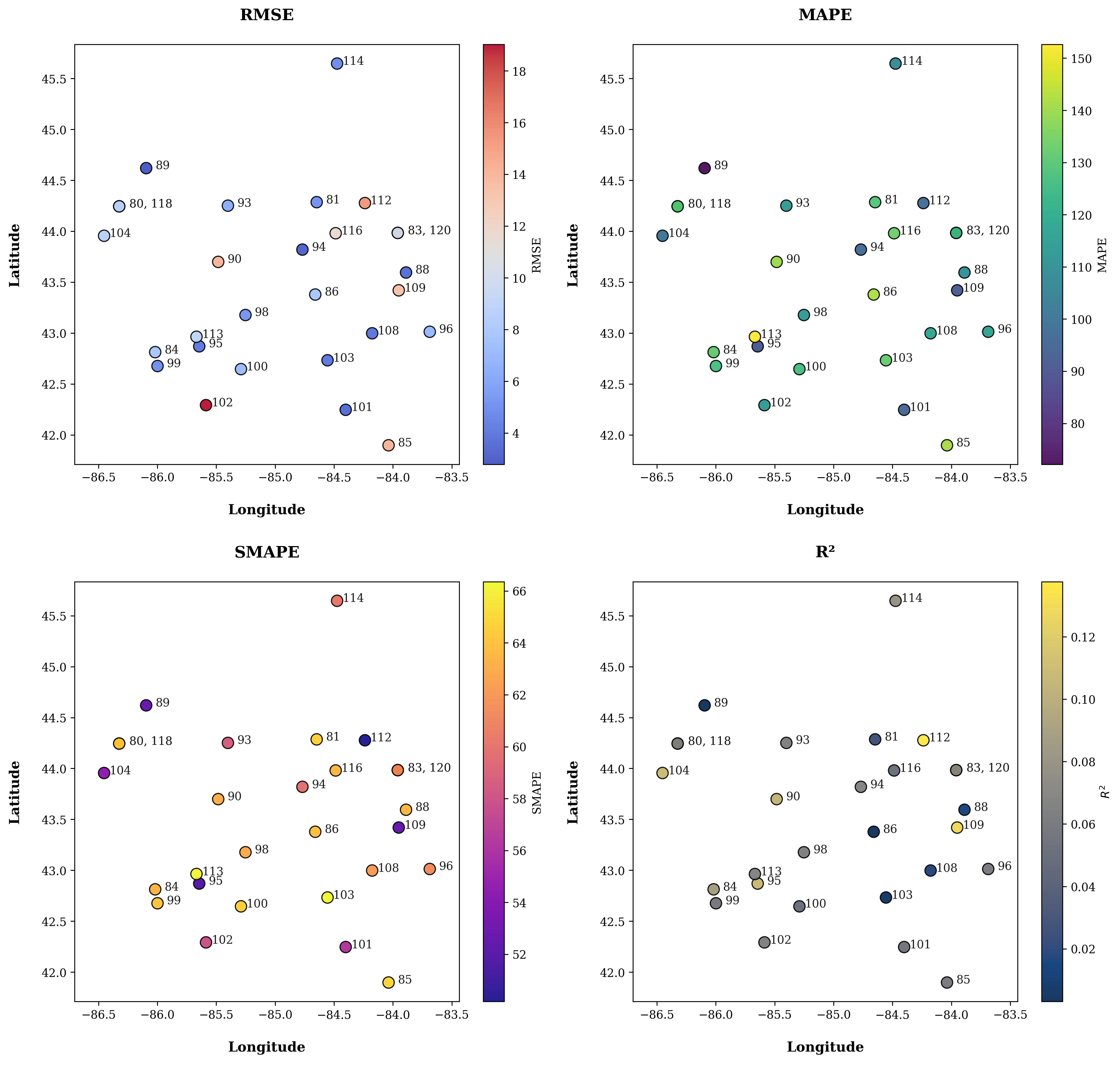}
           \caption{Outage Prediction metrics of Feb 23, 2021}
             \label{Feb24predictionmetrics}	     	     
\end{figure}

The bar chart in figure \ref{Feb24prediction} shows the forecast versus actual power outages of February 24, 2021, trained on February 20, 21, 22, and 23 across different regions. The model shows variable performance across these regions—sometimes forecasts and actuals are close to matching, but striking differences remain in other instances. In areas like 80, 85, and 95, actual outages are very near the forecasted values. Contrasting this, areas like 90, 101, and 109 reflect relatively poor performance—the actual outages were much higher compared to the forecasted ones. 
Considering the individual forecasts' performance, February 21, 22, and 23 performed almost the same throughout the study regions. Contrariwise, the forecast of February 20 shows a larger deviation from the real outage data in the case of the poorest model performance regions, like 90 and 102.

\subsection{Comparison to the benchmark models}

\begin{figure}
\centering
           \includegraphics[width=0.5\textwidth]{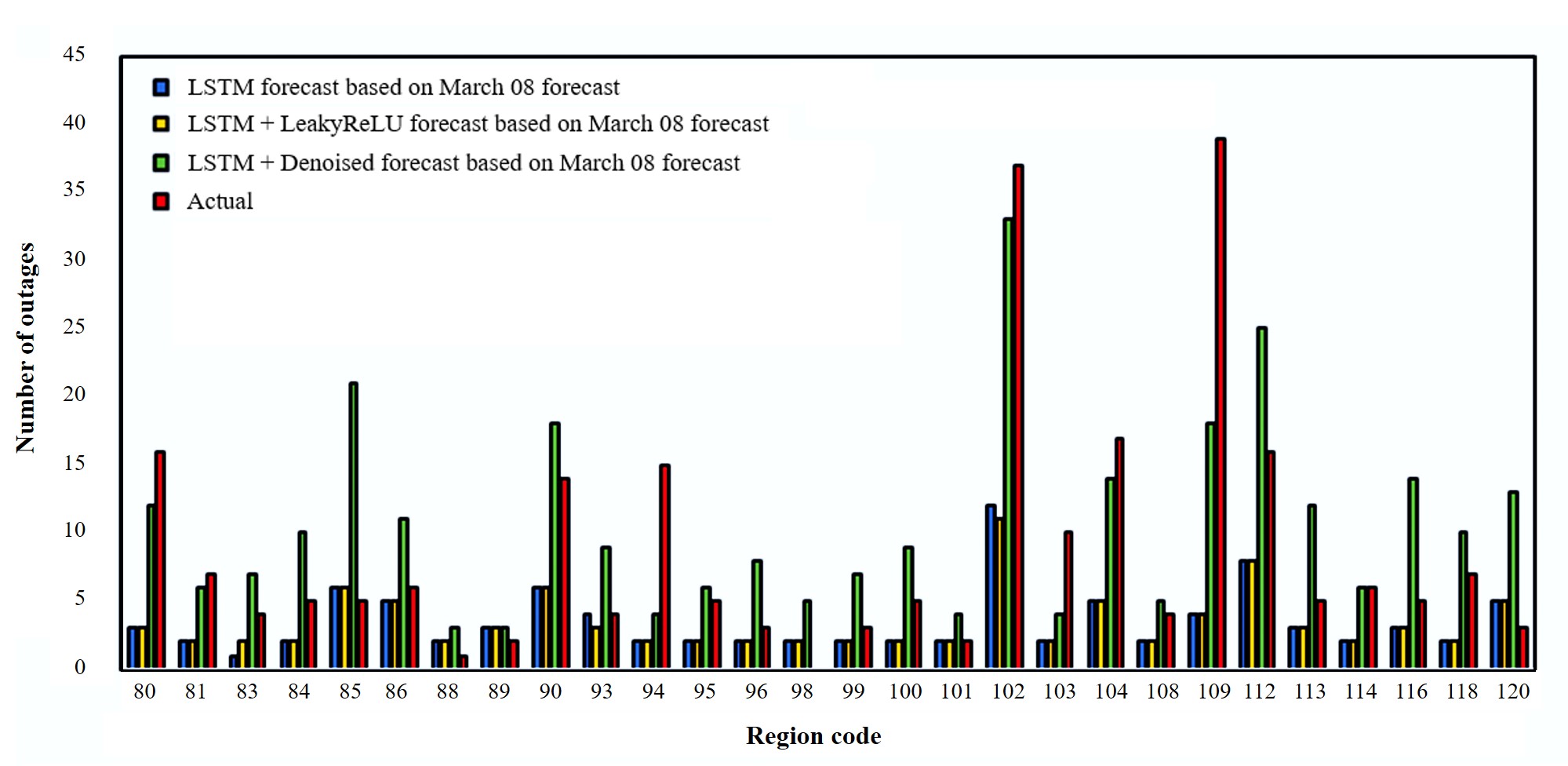}
           \caption{Outage Predictions of Mar 9, 2021.}
             \label{Mar9prediction}	     	     
\end{figure}

\begin{figure}
\centering
           \includegraphics[width=0.5\textwidth]{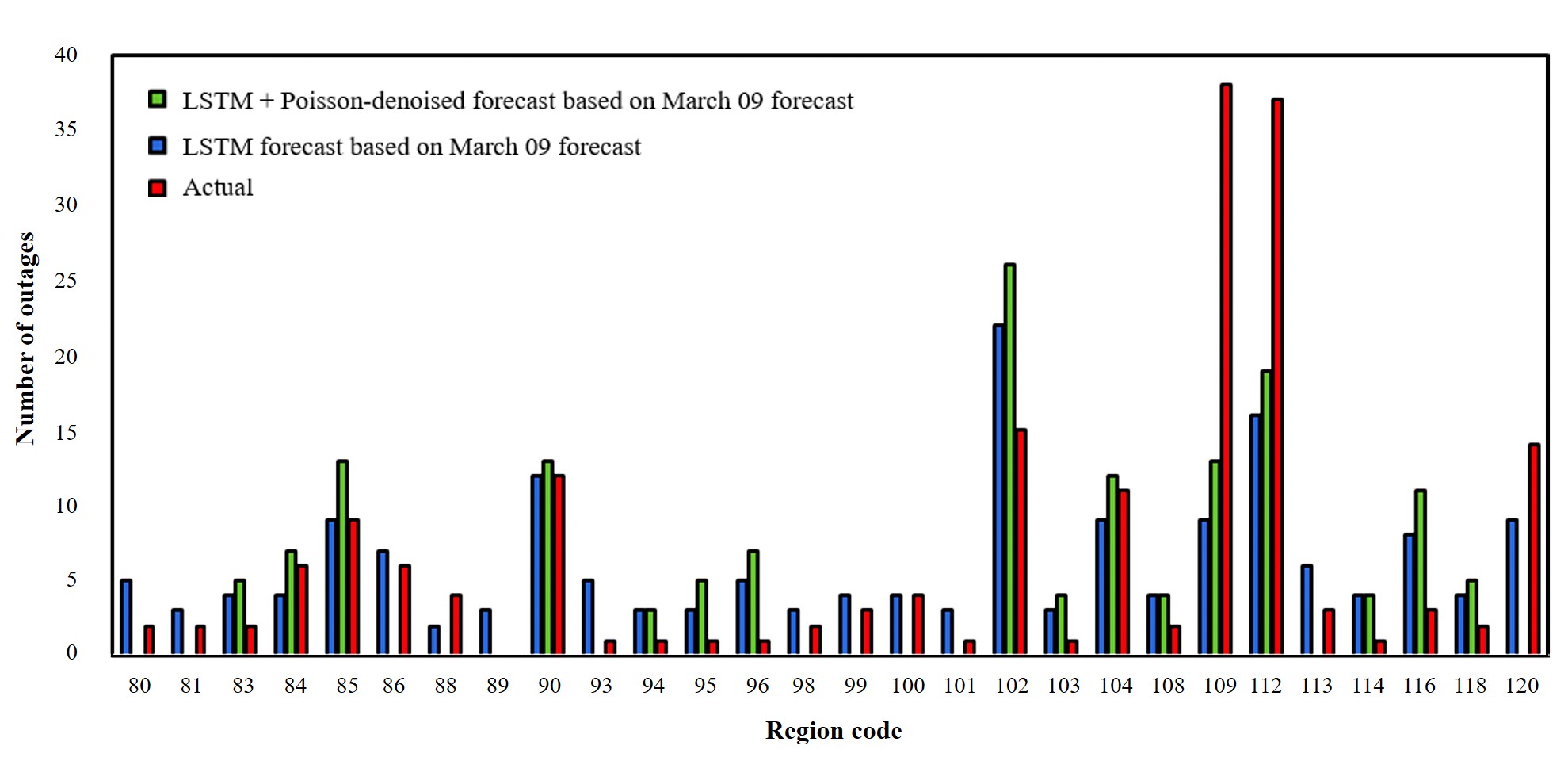}
           \caption{Outage Predictions of Mar 10, 2021}
             \label{Mar10prediction}	     	     
\end{figure}

\begin{itemize}
\item \textbf{LSTM}

    The standalone LSTM model (blue bars in figure \ref{Mar9prediction}) provides the forecast of outage numbers in different regions based on available data from March 8. From this chart, we can identify that the LSTM model tends to underpredict for some areas, like regions 80, 94, 102, and 109, as in reality, the number of outages is more than what the LSTM has predicted. But the rest of the regions of LSTM predictions, like 85, 93, 99, 101, and 113, are almost near to the actual ones, hence proving to be good in those respective areas. 
    Overall, while the LSTM model captures some of the outage patterns, it struggles with higher or more volatile levels of outages across regions.\\
    
    \item \textbf{LSTM+LeakyReLU}
    
    The yellow bars in figure \ref{Mar9prediction} represent the model LSTM + LeakyReLU, which incorporates the LeakyReLU activation function into the LSTM framework to avoid the problem of vanishing gradients faced in deep networks. In general, its forecast shows a similar performance to the basic model LSTM, though for some regions, such as 83, 93, and 102, it forecasts differently. However, this also includes large underestimates of regions such as 80, 94, and 102.  \\
    
    \item \textbf{LSTM+Denoised}
    
    The LSTM + Denoised model, represented by the green bars in figure \ref{Mar9prediction}, includes one step of denoising within the basic LSTM framework, which in turn allows the model to filter out the noise from the data and focus more on the meaningful patterns. As shown by the chart above, this model does much better in capturing high outage numbers in regions like 80, 90, 102, and 104 for which the predictions are closer to the true values. In many places, like regions 81, 95, and 102, this model shows better alignment with actual data compared to the basic LSTM and the LSTM + LeakyReLU model. The denoising approach seems to improve robustness in general for areas with increased variability of the outage data.\\

    Among the above-mentioned models, LSTM + Denoised performs the best in general, especially for regions with heavier outages. The basic LSTM and LSTM + LeakyReLU models also show some improvement but are not as good as their denoised versions in high-outage regions.\\
    
    \item \textbf{LSTM vs. LSTM+ Poisson-Denoised Forecasts}
    
    Figure \ref{Mar10prediction} compares LSTM and LSTM+Poisson-Denoised forecasts for the day of March 10, 2021, using data from March 9. The basic LSTM performed well in some areas, like 85 and 90, where it obtained an approximated value almost equal to the true value. The LSTM+Poisson-denoised model outperforms the basic LSTM in the regions with higher outage values. For example, in area 84, its estimate is much closer to the actual count than that of the basic LSTM model. In the high-outage areas of 109 and 112, it has considerably better predictions. Overall, the performance for the LSTM + Poisson-denoised model is slightly better in high-outage areas compared to that of the basic LSTM, whose best performance areas are where the outages are small.\\
\end{itemize}






\section{Conclusion}
Across all evaluated metrics, including time series distribution of outage frequency and marginal daily occurrences, the hybrid outage prediction model demonstrated significant improvements over standard event-wise classification and regression methods. By integrating Poisson regression, Seq2seq architecture, and PCA components, the model outperformed the baseline LSTM in predicting the spread of outage distributions and total storm-related outages, particularly capturing large peaks missed by other models. The hybrid model's assumption of a Poisson distribution effectively captured noise and highlighted essential correlations between features and outage counts, enabling better predictive accuracy. However, challenges remain, including contamination from changes in power infrastructure, long-term climatic and economic trends, and inherent inaccuracies in weather forecasts. Addressing these limitations could involve using larger datasets spanning multiple decades, accounting for uncontrollable variables, and leveraging ensemble numerical weather predictions to reduce forecast variability. Future work should explore probabilistic outage forecasts to better quantify predictive uncertainties, ultimately improving resource allocation, reducing restoration costs, and minimizing outage durations.




\end{document}